\documentclass[lettersize,journal]{IEEEtran}

\usepackage{amsmath,amssymb,amsfonts}
\usepackage{textcomp}

\usepackage{graphicx}
\usepackage{tikz}
\usepackage{xcolor}

\usepackage{subcaption} 

\usepackage{booktabs}
\usepackage{multirow}
\usepackage{array}
\usepackage{tabularx}

\usepackage{algorithmic}
\usepackage{stfloats}
\usepackage{url}
\usepackage{verbatim}
\usepackage{comment}
\usepackage{afterpage}
\usepackage{etoolbox}
\usepackage{balance}
\usepackage{cite}

\def\BibTeX{{\rm B\kern-.05em{\sc i\kern-.025em b}\kern-.08em
    T\kern-.1667em\lower.7ex\hbox{E}\kern-.125emX}}

\begin{document}
\title{Simulating Realistic LiDAR Data Under Adverse Weather for Autonomous Vehicles: A Physics-Informed Learning Approach}
\author{Vivek Anand, Bharat Lohani, Rakesh Mishra, and Gaurav Pandey 
\thanks{Vivek Anand and Bharat Lohani are with the Geoinformatics Group, Department of Civil Engineering, Indian Institute of Technology Kanpur, India (email: viveka21@iitk.ac.in, blohani@iitk.ac.in)}%
\thanks{Rakesh Mishra is with the Department of Geodesy and Geomatics Engineering, University of New Brunswick, NB, Canada (email: rakesh.mishra@unb.ca)}%
\thanks{Gaurav Pandey is with Department of Engineering Technology \& Industrial Distribution (ETID) of Texas A\&M University (email:
 gpandey@tamu.edu)}%
}


\maketitle

\begin{abstract}


Accurate LiDAR simulation is crucial for autonomous driving, especially under adverse weather conditions. Existing methods struggle to capture the complex interactions between LiDAR signals and atmospheric phenomena, leading to unrealistic representations. This paper presents a physics-informed learning framework (PICWGAN) for generating realistic LiDAR data under adverse weather conditions. By integrating physics-driven constraints for modeling signal attenuation and geometry-consistent degradations into a physics-informed learning pipeline, the proposed method reduces the sim-to-real gap. Evaluations on real-world datasets (CADC for snow, Boreas for rain) and the VoxelScape dataset show that our approach closely mimics real-world intensity patterns. Quantitative metrics, including MSE, SSIM, KL divergence, and Wasserstein distance, demonstrate statistically consistent intensity distributions. Additionally, models trained on data enhanced by our framework outperform baselines in downstream 3D object detection, achieving performance comparable to models trained on real-world data. These results highlight the effectiveness of the proposed approach in improving the realism of LiDAR data and enabling robust perception under adverse weather conditions.

\end{abstract}


\section{Introduction}
Autonomous systems, such as self-driving cars and drones, have emerged as transformative technologies with the potential to revolutionize transportation, logistics, and many other fields \cite{royo2019overview}. These systems rely on a variety of sensors to perceive their environment, make decisions, and navigate safely. Among these sensors, LiDAR (Light Detection and Ranging) plays a pivotal role by providing high-resolution 3D information about the surroundings \cite{schwarting2018planning}.

LiDAR sensors emit laser pulses and measure their return time to create detailed 3D maps, leading to identifying objects and performing tasks like object detection and path planning \cite{lidar_li_2020}. LiDAR intensity, representing the strength of the return signal, provides crucial information on object characteristics and surfaces. Intensity variations help distinguish materials (e.g., concrete, vegetation), identify road markings, and detect surface changes, aiding robust object detection, classification, and scene understanding, which are essential for safe autonomous navigation \cite{wang2018pointseg}\cite{wang2019scnet}

Adverse weather conditions, such as rain and snow, degrade LiDAR performance by scattering and attenuating laser pulses, leading to both intensity distortions and geometric degradations in the point cloud. These effects alter intensity values and introduce noise and point drops, reducing the reliability of LiDAR-based systems \cite{Aw_SurveyOL}. Snowflakes and raindrops can produce spurious returns, while water accumulation further weakens signals \cite{hahner2022lidar}. Such degradations pose significant challenges for autonomous systems in accurate object detection and classification, impacting safety and robustness. To address this, training and validating LiDAR perception algorithms on large datasets that reflect adverse weather conditions is essential \cite{Aw_SurveyOL}. However, collecting such data is costly and time-consuming, making simulation a practical alternative for generating weather-affected LiDAR datasets \cite{8569907}.

\begin{figure*}[!t]
    \centering
    
    \includegraphics[width=0.90\linewidth]{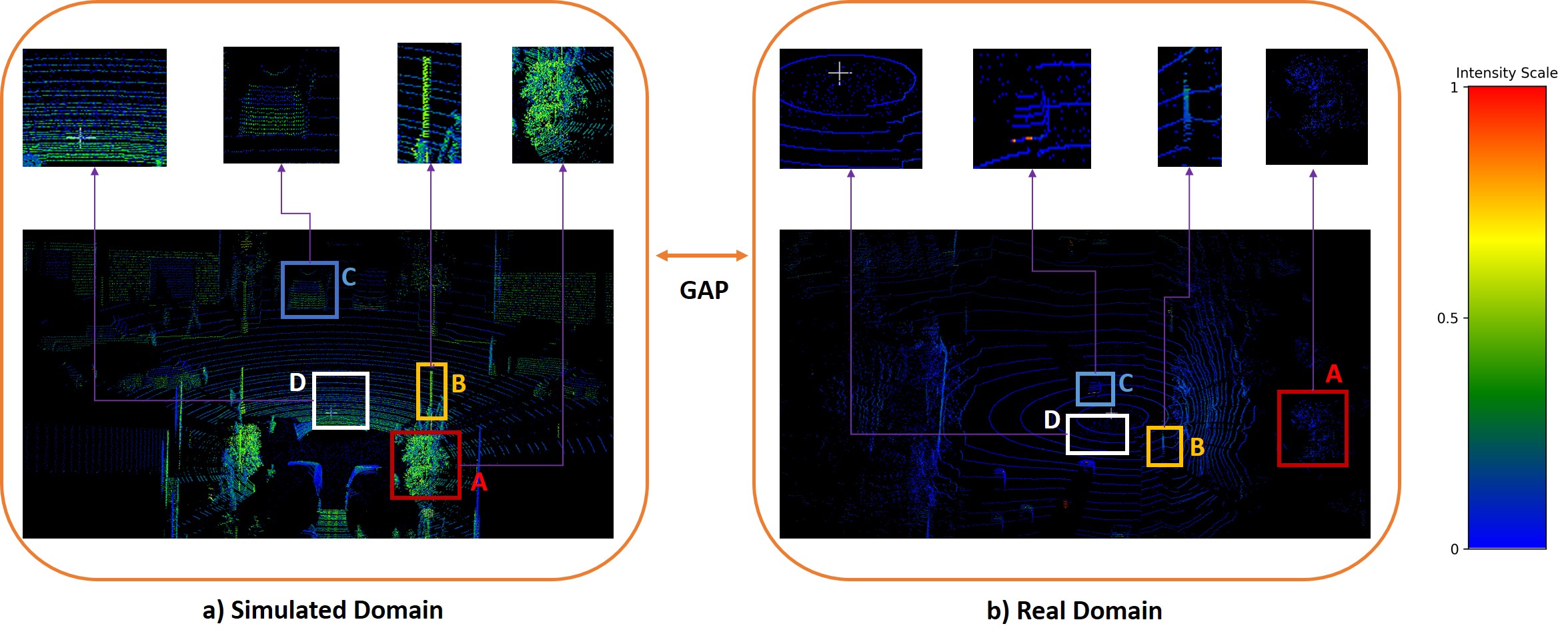}
    
    \caption{\textbf{Simulation-to-Real Gap:} LiDAR point cloud with intensity information: a) Simulated Domain—VoxelScape data augmented with snow using a physics-based method \cite{lisa_kilic2021lidar}; b) Real Domain—CADC data collected in snowy conditions. Objects marked with A (vegetation), B (metallic pole), C (car), and D (road surface). The significant difference in intensity distribution arises because the physics-based simulation fails to capture intensity attenuation due to adverse weather conditions like snow.} 
    \label{fig:sim2real_gap}
\end{figure*}

LiDAR intensity is influenced by factors such as range, incidence angle, material properties, and atmospheric conditions \cite{anand2024toward}. Physics-based simulation methods aim to model these dependencies, but they struggle with complex environmental interactions, especially in adverse weather like rain and snow \cite{Aw_SurveyOL}. The scattering and attenuation effects of snowflakes and raindrops are difficult to model accurately, leading to oversimplified simulations and a significant sim-to-real gap, as shown in Fig. \ref{fig:sim2real_gap}. This gap stems from stochastic variations in snowflake and raindrop properties, as well as material-dependent surface changes not captured by simulations.


The sim-to-real gap impedes the deployment of learning-based perception algorithms, making it necessary to reduce this sim-to-real gap to train reliable autonomous systems. Learning-based methods like CycleGAN \cite{cyclegan_zhu2017unpaired} offer a promising solution by aligning simulated data with real-world distributions, but they lack utility in tasks that require the preservation of physical meaning, such as LiDAR intensity translation. In this research, we develop a Physics-Informed Cycle-consistent Weather GAN (PICWGAN) framework to enhance the realism of simulated LiDAR data under adverse weather conditions. Unlike standard CycleGANs, which are designed for RGB image translation without enforcing physical consistency, LiDAR signals are fundamentally grounded in physics. We provide a framework that embeds physics-based constraints accounting for range, incidence angle, material reflectance, weather effects (rain and snow), and geometric degradation within a generative framework. This integration is crucial for achieving accurate translation, a task that conventional CycleGANs cannot accomplish due to their lack of physical awareness.

By combining learning-based methods with physics-based constraints, our approach addresses the limitations of traditional simulation methods and reduces the sim-to-real gap. This work represents a significant step toward improving the robustness of LiDAR perception systems under adverse weather conditions, enabling more reliable operation of autonomous systems in a challenging environment.

The key contributions of this study are:
\begin{itemize}
    \item \textbf{Realistic LiDAR Simulation for Adverse Weather:} Developed a hybrid framework that combines physics and learning-based methods to accurately model the effects of adverse weather on LiDAR signals, including both intensity and geometric degradation.

    \item \textbf{Adverse-Weather Data Augmentation Framework:} Established a framework to transform both simulated and real clear weather data into realistic adverse weather data, enabling broader utilization of existing datasets.

    \item \textbf{Real-World Applicability:} Demonstrated the potential of the framework for autonomous systems, improving safety and reliability under adverse weather in downstream tasks.
\end{itemize}

The source code and data supporting the experiments and analyses in this paper are publicly available at https://github.com/voodooed/LBLIS-Adverse-Weather. In addition, we will release KITTI-Snow and KITTI-Rain, augmented versions of the original KITTI dataset with realistic snow and rain effects on LiDAR signals generated using our framework to support research in adverse weather conditions. We further plan to extend the repository with similarly augmented versions of other widely used open-source datasets.

 \section{Related Work}


 \subsection{Physics-based Simulation in Adverse Weather}

Physics-based simulation methods model the interaction between LiDAR signals and environmental factors like rain, snow, and fog to generate realistic LiDAR data. These methods typically involve \cite{rashofer_aw_theo_model}:

\begin{itemize}
    \item \textbf{Theoretical LiDAR Model:}  A mathematical model that describes the relationship between emitted LiDAR pulses, scene reflectivity, and received intensity. This model accounts for factors like attenuation, scattering, and reflection.

    \item \textbf{Environmental Parameters:} Specifying weather conditions such as rain rate, snow density, and extinction coefficient.
\end{itemize}

Rasshofer et al. \cite{rashofer_aw_theo_model} propose a theoretical LiDAR model that accounts for rain, fog, and snow effects on intensity by treating the system as linear and convolving the emitted pulse with the scene response. Yang et al. \cite{Yang2023RealisticRW} focus on rain effects, particularly spray and splash from vehicle wheels, to augment LiDAR data and improve perception. Teufel et al. \cite{Teufel_sim_rain} use physics and empirical models to simulate rain, snow, and fog, emphasizing parameterization for real-world accuracy. Goodin et al. \cite{Goodin_rain} predict LiDAR performance degradation in rain, based on rain rate. Kilic et al. \cite{lisa_kilic2021lidar} introduce a model considering attenuation, scattering, and noise in rain to predict point loss or scattering. Hahner et al. \cite{hahner2022lidar} introduces a model for snowfall and a wet-ground method to estimate noise floor and ground point loss during snow.
 
While physics-based methods offer a principled approach to LiDAR intensity simulation, they often rely on simplifying assumptions and may not capture the full complexity of LiDAR interaction with adverse weather. These limitations can lead to:




\begin{itemize}
\item \textbf{Oversimplification of Real-World Effects:} High variability in weather conditions and their complex interactions with LiDAR can be difficult to model accurately, resulting in oversimplified simulations. 

\item \textbf{Limited Generalizability:} These models require specific parameterization for particular weather conditions and sensor configurations, limiting their adaptability to different scenarios. 

\item \textbf{Inaccurate Attenuation Modeling:} Modeling attenuation from rain and snow can be challenging, leading to unrealistic intensity values in simulations.

\end{itemize}

\subsection{Learning-based Simulation in Adverse Weather}

 

Learning-based methods \cite{anand2025advancing}, though still emerging, offer a promising alternative to physics-based approaches for LiDAR simulation, addressing their limitations by using data-driven models to capture complex interactions and simulate adverse weather effects.

Lee et al. \cite{Lee_gan_trans} propose a GAN-based model that transforms point clouds from sunny to foggy or rainy conditions, achieving results comparable to real-world data but without addressing LiDAR intensity generation in adverse weather. Zhang et al. \cite{ldig_zhang} introduce the L-DIG model for noise reduction and snow point synthesis on LiDAR point clouds, using depth images and custom loss functions for scale and structure consistency, but it does not focus on intensity generation. Zhang et al. \cite{cgan_zhang} present a conditional generative model to integrate realistic weather effects into clear-weather LiDAR datasets, using segmentation maps for robust perception, yet it doesn’t address intensity simulation.
 

Physics-based methods rely on simplistic assumptions that fail to capture the complexity of LiDAR interactions in adverse weather \cite{Lee_gan_trans}. While learning-based methods can model complex, nonlinear interactions, most focus on noise effects, such as point addition or removal. No existing study addresses LiDAR intensity simulation in adverse weather. This gap is significant, as LiDAR intensity is crucial for object characterization and classification. This work bridges the gap by proposing a hybrid approach that combines physics-based principles with learning-based frameworks to simulate LiDAR intensity under adverse weather conditions.

\section{Methodology}

The objective of this work is to develop a physics-informed, learning-based approach for simulating LiDAR intensity under adverse weather conditions such as rain and snow. Conventional data-driven image translation methods often introduce artifacts, which is particularly problematic given the physical significance of LiDAR intensity. To address this, we incorporate physics-based constraints into the learning process, ensuring better alignment with real-world intensity distributions while also accounting for weather-induced geometric degradation. This formulation mitigates artifacts and bridges the sim-to-real gap, improving intensity accuracy by modeling factors such as incidence angle, material reflectance, range, and weather-induced attenuation. The resulting realism supports robust training and validation of LiDAR perception algorithms for autonomous systems.

\subsection{\textbf{Data}}
This study utilizes datasets from two distinct domains:
\begin{enumerate}
    \item Real-world adverse weather domain:
    \begin{itemize}
        \item Canadian Adverse Driving Conditions (CADC) dataset \cite{cadc_pitropov2021canadian}, captured under snowy weather conditions.
        \item Boreas dataset \cite{boreas_burnett2023boreas}, captured under rainy weather conditions.
    \end{itemize}
    \item Simulated domain:
    \begin{itemize}
        \item VoxelScape dataset \cite{voxelscape}, which provides simulated LiDAR data under clear weather conditions.
    \end{itemize}
\end{enumerate}

\begin{figure*}[]
\centerline{\includegraphics[width=0.80\textwidth,keepaspectratio]{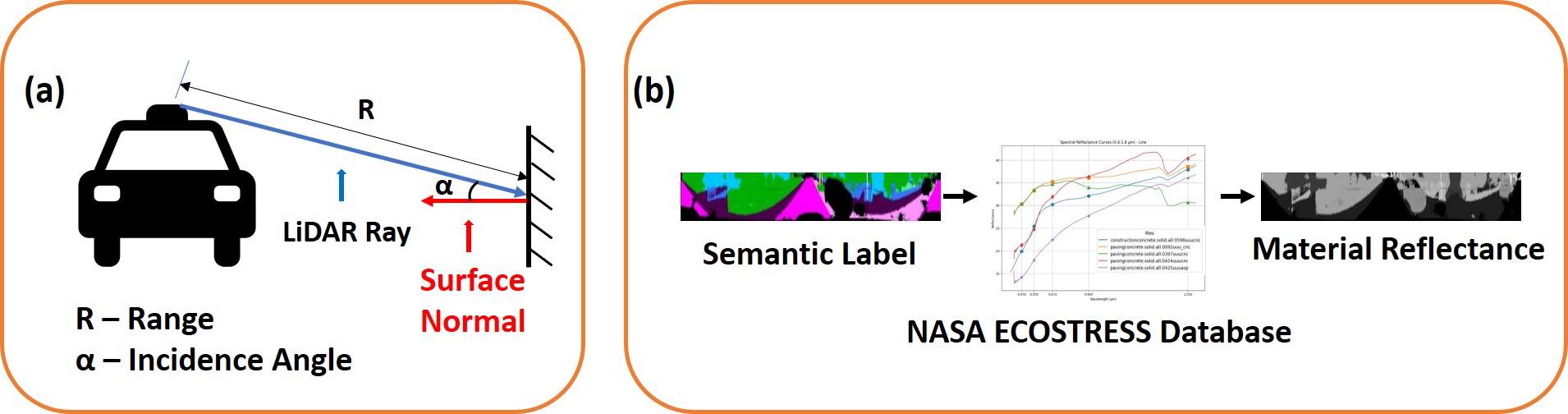}}
\caption{\textbf{Physical Modalities:}\textbf{(a)} The range is calculated using the Euclidean distance between the LiDAR sensor and the point, while the incidence angle is determined using the dot product of the LiDAR ray direction vector and surface normal vectors. \textbf{(b)} Material reflectance is obtained by mapping semantic labels in the LiDAR data to spectral reflectance values from NASA's ECOSTRESS Library, based on identified materials.}
\label{fig-modalities}
\end{figure*}

\subsubsection*{\textbf{Augmenting Simulated Data with Adverse Weather Effects}}

Accurately simulating LiDAR under adverse weather conditions (snow and rain) requires modeling both intensity variations and changes in point cloud geometry. To this end, we augment simulated clear-weather data with noise points and point drop effects using a physics-based approach, realized through a hybrid Monte Carlo formulation \cite{lisa_kilic2021lidar} to capture the impact of adverse weather on the spatial structure of LiDAR point clouds.

\begin{itemize} 

\item \textbf{Noise point simulation:} Atmospheric scattering from particles (e.g., rain droplets, snowflakes) introduces spurious returns. We model this by placing virtual scatterers based on physical distributions and comparing their backscattered power with that of true scene objects. 

\item \textbf{Point drop simulation:} Scattering and absorption attenuate laser beams, leading to point loss when the returned intensity falls below the sensor noise threshold. Using Mie theory \cite{lisa_kilic2021lidar}, we compute scattering efficiencies across particle sizes and distributions to model attenuation effects. 

\end{itemize}

The approach incorporates LiDAR-specific parameters (e.g., range and beam divergence) along with real-world weather conditions (e.g., rain rate) to estimate extinction coefficients, capturing signal loss due to scattering and absorption. This physics-based augmentation models geometric degradations arising from noise and point drops, enabling realistic representation of spatial distortions under adverse weather.

By decoupling geometric effects from intensity representation, the proposed PICWGAN framework operates on geometrically consistent data, allowing it to focus on learning LiDAR intensity variations induced by environmental factors. This results in data that is both spatially coherent and physically accurate in intensity, significantly enhancing the realism and applicability of the simulated LiDAR data.



In this work, LiDAR point cloud data is transformed into spherical projections to create 2D images that represent the 3D environment \cite{sph_proj_li2016vehicle}. This approach integrates diverse input features—such as range, incidence angle, and material reflectance—encoded as pixel values \cite{anand2025advancing}. The resulting representation provides rich spatial and material information, enabling the model to predict intensity with greater precision by leveraging the synergy of these modalities. The physical modalities derived from raw LiDAR data, shown in Fig. \ref{fig-modalities}, serve as inputs to the architecture. We use the following modalities as input to our framework:
\begin{itemize}
    \item \textbf{Range:} For each point in the point cloud, the range to the LiDAR sensor is calculated using the Euclidean distance formula. This crucial parameter directly influences the intensity of the returned signal, as signals attenuate with increasing distance. By incorporating range information, the model can learn to simulate the distance-dependent attenuation of the LiDAR signal.

    \item \textbf{Incidence Angle:}
   The incidence angle, defined as the angle between the incoming laser beam and the surface normal at each point, critically affects reflected light intensity. Lower incidence angles result in higher reflected intensity and vice versa. To compute this, the local surface orientation (normal vector) is estimated and aligned to face the sensor. The LiDAR ray direction vector for each point is calculated, and the incidence angle is obtained using the dot product of the normal and direction vectors. This enables the model to account for surface reflectivity variations based on the angle of incidence.

    \item \textbf{Material Reflectance:} Material reflectance, which describes the ratio of reflected energy to incident energy at specific wavelengths, is integrated into the architecture using data from NASA’s ECOSTRESS spectral library \cite{ECOSTRESS}. This library includes over 3,400 reflectance profiles for various natural and man-made materials. Reflectance values from the library are mapped to semantic labels within the LiDAR dataset, based on the sensor's wavelength. Each object in the scene is associated with a material based on common knowledge (e.g., asphalt for roads, wood for trees). This mapping enables the model to account for reflectivity variations across different materials, leading to more realistic intensity predictions. 

This comprehensive set of input features provides the PICWGAN framework with a rich representation of the scene geometry, material properties, and sensor-object interactions, enabling it to learn complex mappings between simulated and real-world LiDAR intensity distributions.
\end{itemize}

\subsection{\textbf{Physics-Informed Cycle-Consistent Weather GAN (PICWGAN) Architecture}}

Standard image translation frameworks lack physical awareness, making them unsuitable for LiDAR intensity modeling. To address this, we develop a physics-informed learning framework that embeds physics-based constraints within a modified generative framework. Specifically, we incorporate the physics of LiDAR intensity attenuation under adverse weather conditions directly into the generator as a guiding constraint. The intensity of the returned LiDAR signal in such conditions is governed by the following factors \cite{rashofer_aw_theo_model}:

\begin{itemize}

\item \textbf{Surface Interaction and Backscatter Modeling:}  
The returned LiDAR intensity is governed by the interaction between the emitted beam and the target surface, characterized by material reflectance, incidence geometry, and propagation range. Under a Lambertian assumption, the backscattered intensity can be expressed as \cite{voxelscape}:
\begin{equation}
I_{\text{phy}} = \frac{\rho \cdot \cos\theta}{R},
\end{equation}
where $\rho$ denotes the material reflectance, $\theta$ is the angle between the incident beam and the surface normal, and $R$ is the radial distance from the sensor to the target.

\item \textbf{Atmospheric Attenuation under Adverse Weather:}  
During propagation through adverse weather, the LiDAR beam undergoes attenuation due to scattering and absorption by atmospheric particles. This effect is modeled via the Beer--Lambert formulation \cite{lisa_kilic2021lidar}, yielding
\begin{equation}
I_{\text{AW}} = I_{\text{phy}} \exp\left(-2 \int_{0}^{R} \alpha(s)\, ds \right),
\end{equation}
where $\alpha(s)$ is the path-dependent extinction coefficient. Assuming a homogeneous medium, this reduces to
\begin{equation}
I_{\text{AW}} = I_{\text{phy}} \, e^{-2\alpha R},
\end{equation}
with $\alpha$ parameterized as $\alpha = 1.45 R_r^{0.64}$, where $R_r$ denotes the precipitation rate in mm/hr (30 mm/hr in this work).

\end{itemize}

The resulting formulation of $I_{\text{AW}}$ is integrated as a physics-based constraint within the generative process, enforcing consistency with radiative transport principles and ensuring physically plausible intensity predictions under adverse weather conditions.


\subsection{\textbf{Training PICWGAN}}

\begin{figure*}[]
  \centering
  \includegraphics[width=0.95 \textwidth,keepaspectratio]{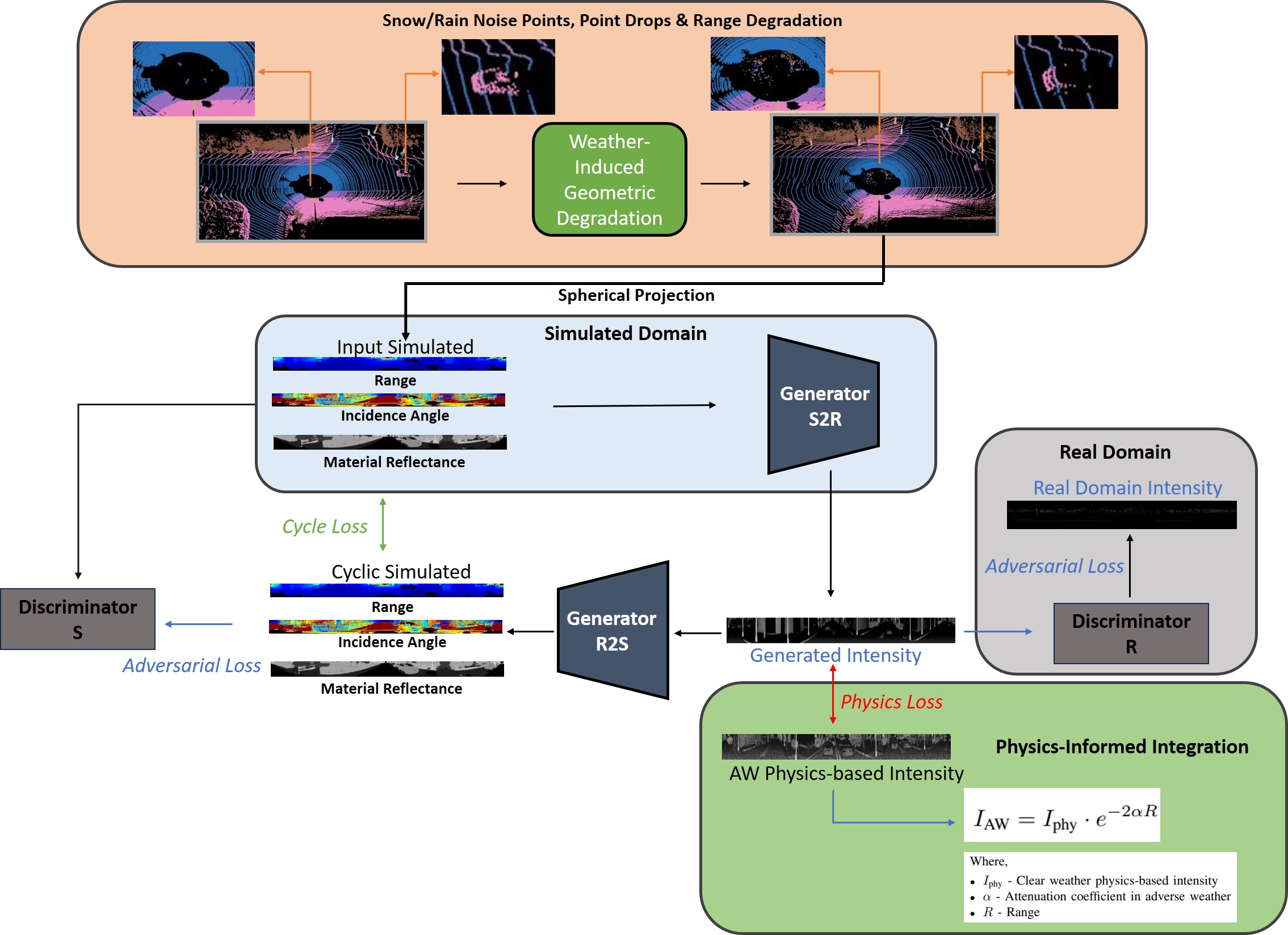}

  \caption{\textbf{PICWGAN Training framework:} LiDAR point clouds, projected onto a spherical surface, are input into our proposed PICWGAN architecture to generate LiDAR intensity under adverse weather conditions.}
  \label{fig:training_pipeline}
\end{figure*}

\begin{figure*}[]
  \centering
  \includegraphics[width=0.90 \textwidth,keepaspectratio]{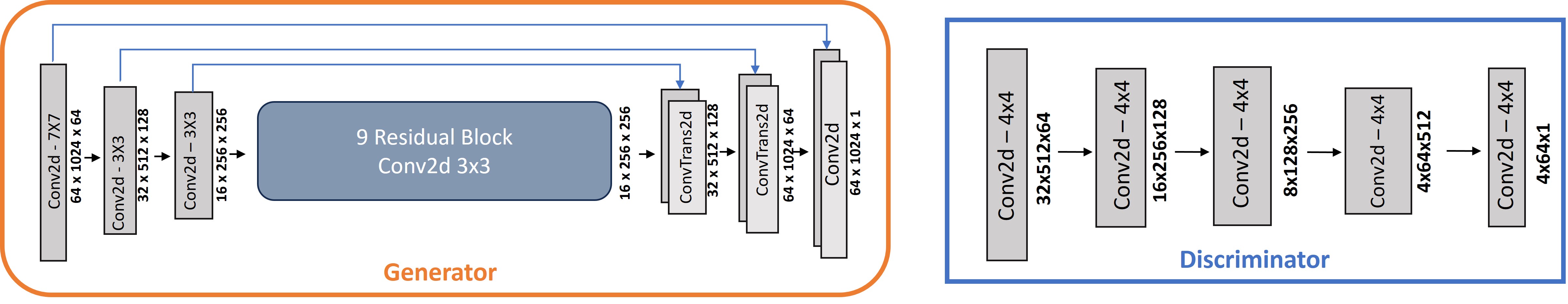}

  \caption{\textbf{Architecture:} Generator and Discriminator architecture.}
  \label{fig:gen_disc}
\end{figure*}

The training process of our PICWGAN framework, which is depicted in Fig. \ref{fig:training_pipeline}, is guided by a carefully designed combination of loss functions. Each loss function plays a distinct role in ensuring both realism and adherence to physical principles. The training process utilizes 5000 frames from both the simulated and real domains for each weather condition (snow and rain) to effectively learn the complex mappings between the domains. These loss functions are described below:

\textbf{Adversarial Loss:} 
    This component, derived from the standard CycleGAN framework, encourages the generator to produce synthetic intensity images that are indistinguishable from real-world data. This is achieved by employing adversarial training, where the generator aims to deceive a discriminator network that attempts to distinguish between real and generated intensities. The adversarial loss is represented as:
    \begin{equation}
\begin{aligned}
    L_{adv}(G_{R2S}, D_S) &= \mathbb{E}_{s \sim p_{\text{data}}(s)} [\log D_S(s)] \\
    &\quad + \mathbb{E}_{r \sim p_{\text{data}}(r)} [\log (1-D_S(G_{R2S}(r)))]
\end{aligned}
\end{equation}

\textbf{Cycle-Consistency Loss:} 
    This loss ensures that the generator can accurately translate images between the real and simulated domains and back again, maintaining the integrity of the original data. It addresses the need for the generated data to maintain consistency with the initial input, creating a reliable, cyclical process. It is given by:
  \begin{equation}
\begin{aligned}
    L_{cycle}(G_{R2S}, G_{S2R}) &= \mathbb{E}_{r \sim p_{\text{data}}(r)} [||r - G_{S2R}(G_{R2S}(r))||_1] \\
    &\quad + \mathbb{E}_{s \sim p_{\text{data}}(s)} [||s - G_{R2S}(G_{S2R}(s))||_1]
\end{aligned}
\end{equation}

where \(r\) represents intensity values from  the real adverse weather domain \(R\), \(s\) represents intensity values from the simulated domain \(S\), \(G_{R2S}\) is the generator mapping from \(R\) to \(S\), \(G_{S2R}\) is the generator mapping from \(S\) to \(R\), and \(D_S\) is the discriminator for domain \(S\).

\textbf{Physics Loss:} 
To enforce consistency with the underlying physical principles governing LiDAR intensity under adverse weather, we introduce a physics-based regularization term. This loss penalizes deviations between the intensities generated by the model and those predicted by the physics-driven adverse weather formulation. Consequently, the model is guided to respect the intrinsic dependencies on material reflectance, incidence angle, range, and atmospheric attenuation.

Formally, the physics loss is defined as:
\begin{equation}
L_{Phy}(G_{S2R}, I_{Phy}) = \frac{1}{N} \sum_{i=1}^{N} \left\lVert I_{\text{gen}}^{(i)} - I_{\text{AW}}^{(i)} \right\rVert_1,
\end{equation}
where $I_{\text{gen}}^{(i)}$ denotes the generated intensity at point $i$, and $I_{\text{AW}}^{(i)}$ is the corresponding intensity computed from the physics-based adverse weather model. 

\noindent The overall loss function is a weighted combination of the individual loss terms, providing a balanced approach during the training process:

\begin{equation}
\begin{aligned}
    L_{total} &=  L_{adv}(G_{S2R}, D_R) +  L_{adv}(G_{R2S}, D_S) \\
    &\quad + \lambda_{cycle} \cdot L_{cycle}(G_{R2S}, G_{S2R}) \\
    &\quad + \lambda_{physics} \cdot L_{Phy}(G_{S2R}, I_{AW})
\end{aligned}
\end{equation} 

where $\lambda_{\text{cycle}}$ and $\lambda_{\text{physics}}$ are hyperparameters that balance the contributions of cycle-consistency and physics-based constraints.

The model is optimized using the Adam optimizer with a learning rate of $0.0001$. The weights for the cycle-consistency and physics losses ($\lambda_{\text{cycle}}$ and $\lambda_{\text{physics}}$) are set to $10$ to ensure an appropriate balance between realistic domain translation and physical fidelity.


By optimizing the combined loss function, the PICWGAN architecture generates LiDAR intensity values that visually resemble real-world data while adhering to the physical principles of LiDAR signal propagation and interaction under adverse weather. This ensures the simulated data is both visually plausible and physically accurate, enhancing its reliability and utility for downstream applications.

\section{Results and Discussion}

This section presents the results of the proposed PICWGAN framework for simulating LiDAR under adverse weather conditions (snow and rain). Geometric degradations are incorporated via a physics-based formulation prior to physics-informed learning, ensuring spatial consistency. The model is evaluated through qualitative and quantitative analyses, with a focus on intensity simulation as the primary learned component. We further validate the realism of the simulated data on a downstream 3D object detection task.

\subsection{\textbf{Qualitative Results}}

\begin{figure*}[!t]
    \centering
    
    \includegraphics[width=0.8\linewidth]{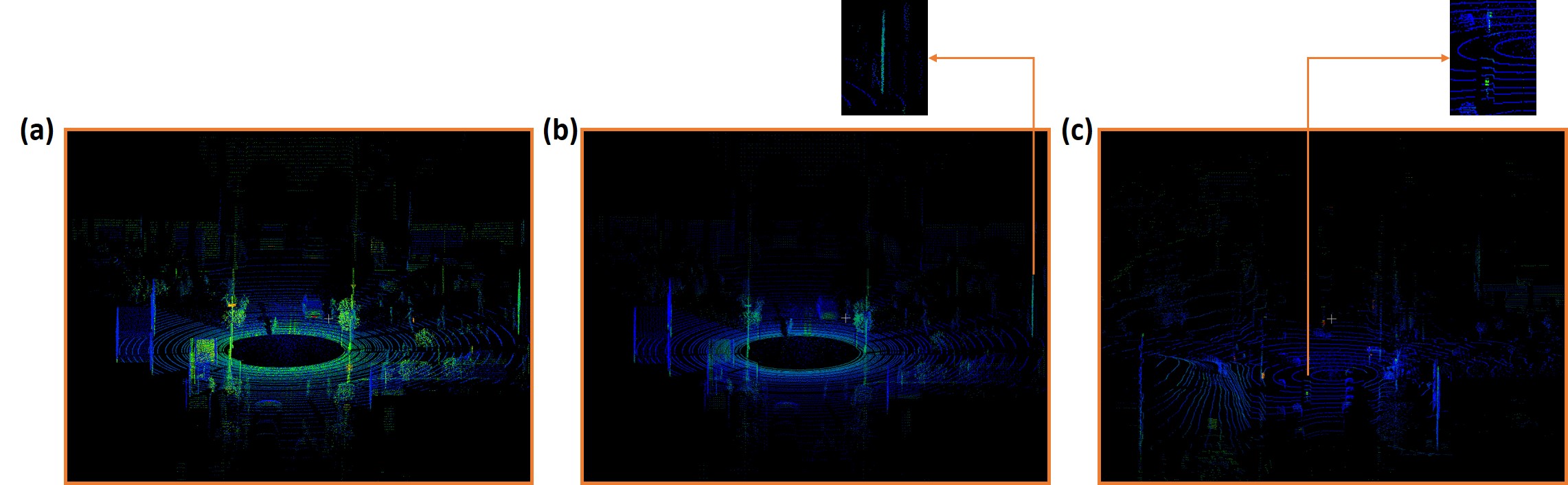}
    
    \caption{\textbf{Qualitative Analysis - 3D:} Visual assessment of the back-projected 3D point cloud with intensity information by our PICWGAN architecture trained on the snow weather CADC dataset. (a) Physics-based simulated snow-weather LiDAR point cloud from the VoxelScape dataset; (b) Same point cloud (from a) after domain adaptation by our PICWGAN; (c) Ground truth LiDAR point cloud from the CADC dataset used for training the PICWGAN.}
    \label{fig:3d_snow}
\end{figure*}

\begin{figure*}[!t]
    \centering
    
    \includegraphics[width=0.8\linewidth]{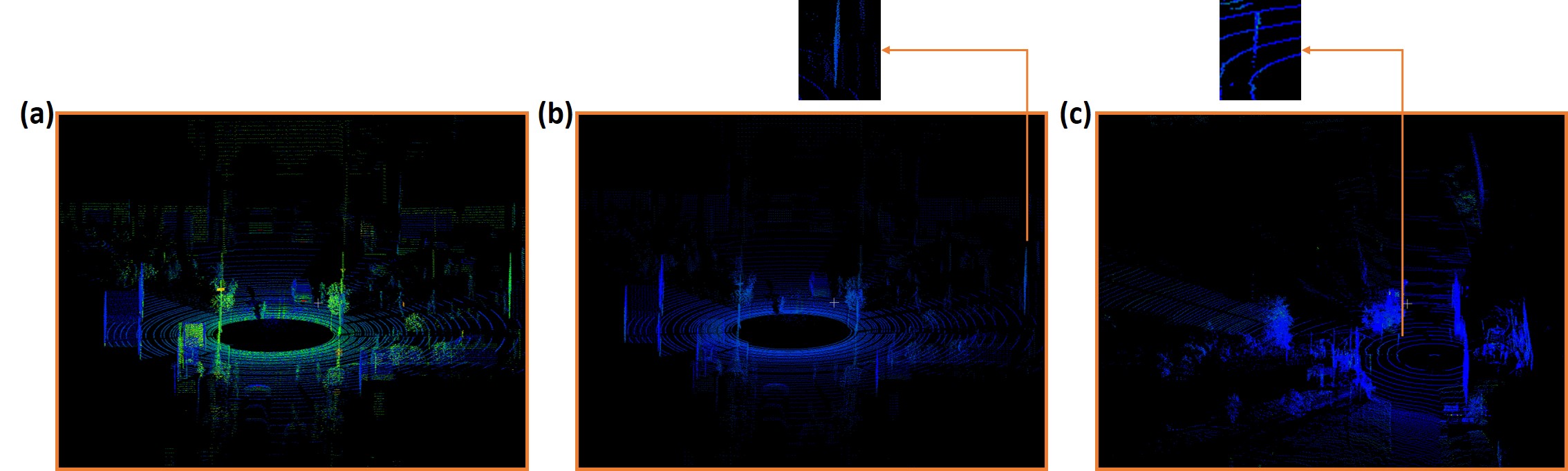}
    
    \caption{\textbf{Qualitative Analysis - 3D:} Visual assessment of the back-projected 3D point cloud with intensity information by our PICWGAN architecture trained on the rain weather Boreas dataset. (a) Physics-based simulated rain-weather LiDAR point cloud from the VoxelScape dataset; (b) Same point cloud (from a) after domain adaptation by our PICWGAN; (c) Ground truth LiDAR point cloud from the Boreas dataset used for training the PICWGAN.}
    \label{fig:3d_rain}
\end{figure*}

The PICWGAN framework generates realistic LiDAR intensity distributions under adverse weather conditions that closely match real-world patterns. As shown in Figs. \ref{fig:3d_snow} and \ref{fig:3d_rain}, the resulting 3D point clouds exhibit characteristic attenuation effects of snow and rain. Regions with reduced intensity correspond to areas where LiDAR signals are weakened due to scattering and absorption by atmospheric particles, leading to diminished returns in affected regions.

The generated point clouds retain physically consistent intensity variations, reflecting the impact of adverse weather on both signal propagation and surface interactions. A direct comparison of Fig. \ref{fig:3d_snow}b and Fig. \ref{fig:3d_rain}b reveals distinct intensity patterns arising from the differing interactions of snow and rain with LiDAR sensors and scene materials. In particular, in rain-affected scenes, water accumulation on highly reflective surfaces such as metal poles and traffic signs results in increased absorption and scattering, producing lower intensity returns compared to snow.

These variations are further influenced by the underlying datasets (CADC for snow and Boreas for rain), which capture real-world differences in sensor response under adverse weather. Overall, these visual results demonstrate the ability of PICWGAN to generate geometrically consistent and physically plausible LiDAR data, while generalizing effectively across diverse weather conditions for reliable 3D perception.

\subsection{\textbf{Quantitative Results}}

To enable a quantitative evaluation of the performance of our PICWGAN model in adverse weather conditions, a same-domain validation strategy was employed. This was necessary because the direct evaluation of real-to-simulated domain intensity transfer is hindered by the fundamental nature of our data: the input and output pairs consist of simulated and real-world lidar data, respectively, inherently lacking direct, paired correspondences. Therefore, we conducted experiments using the CADC (snow) and Boreas (rain) datasets, where ground-truth intensity data were readily available. Specifically, the model was trained on 5,000 frames from each dataset. Notably, the target intensity images were drawn from a distinct, non-overlapping set of 5,000 frames within the same respective datasets, creating an unpaired input-target scenario. This approach allowed us to leverage the existing ground truth intensity data for robust quantitative analysis of the model's ability to generate accurate intensity distributions. The accuracy of generated intensities was assessed by comparing their probability distribution functions (PDFs) and error histograms with ground-truth data to ensure statistical alignment. Metrics such as Mean Squared Error (MSE), Structural Similarity Index (SSIM), Kullback-Leibler (KL) divergence, and Wasserstein Distance were used to quantify deviations between predicted and ground truth intensity distributions.

All validations in the following paragraphs were performed on 2,500 test frames, with the results representing the combined analysis across these frames to ensure statistical significance. The evaluation confirmed that the PICWGAN effectively reproduces realistic intensity distributions, enhancing the realism and accuracy of simulated LiDAR intensity data in adverse weather.

\subsubsection{\textbf{Probability Distribution Function (PDF)}}
\begin{figure*}[h]
\centerline{\includegraphics[width=0.65\textwidth,keepaspectratio]{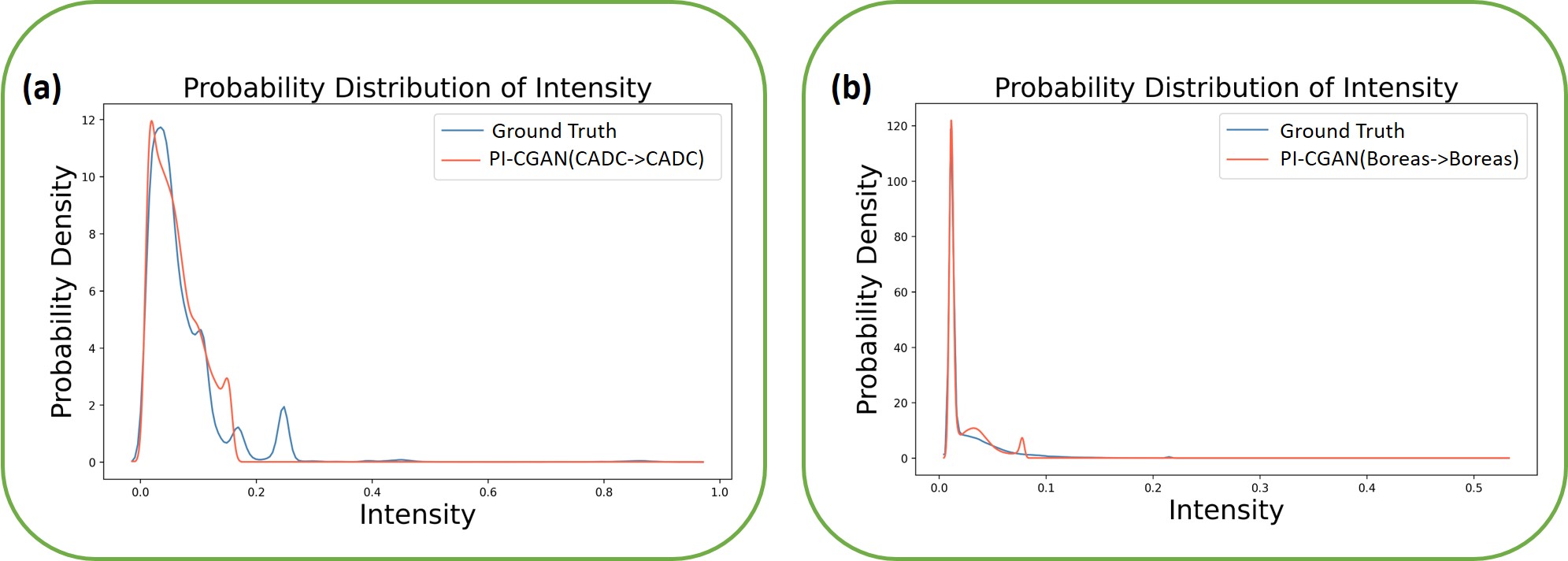}}
\caption{\textbf{Probability Distribution Function} for Same-Domain LiDAR Intensity Transfer on a) Snow Weather CADC dataset b) Rain Weather Boreas dataset between the Ground Truth and our PICWGAN Generated Intensity: PI-CGAN(\text{CADC}~$\rightarrow$~\text{CADC}) and PI-CGAN(\text{Boreas}~$\rightarrow$~\text{Boreas}).} 
\label{fig-pdf}
\end{figure*}

    The PDF is a crucial metric for comparing the distribution of the generated LiDAR intensity values with the ground truth intensities. For both snow and rain, the matching PDFs, as shown in Fig. \ref{fig-pdf}, indicate that the simulated intensity closely replicates the statistical properties of the real-world data. This shows that the model effectively captured the underlying intensity distribution affected by different weather conditions, which is critical for generating realistic data for perception tasks.

\subsubsection{\textbf{Error Histogram}}
\begin{figure*}[h]
\centerline{\includegraphics[width=0.65\textwidth,keepaspectratio]{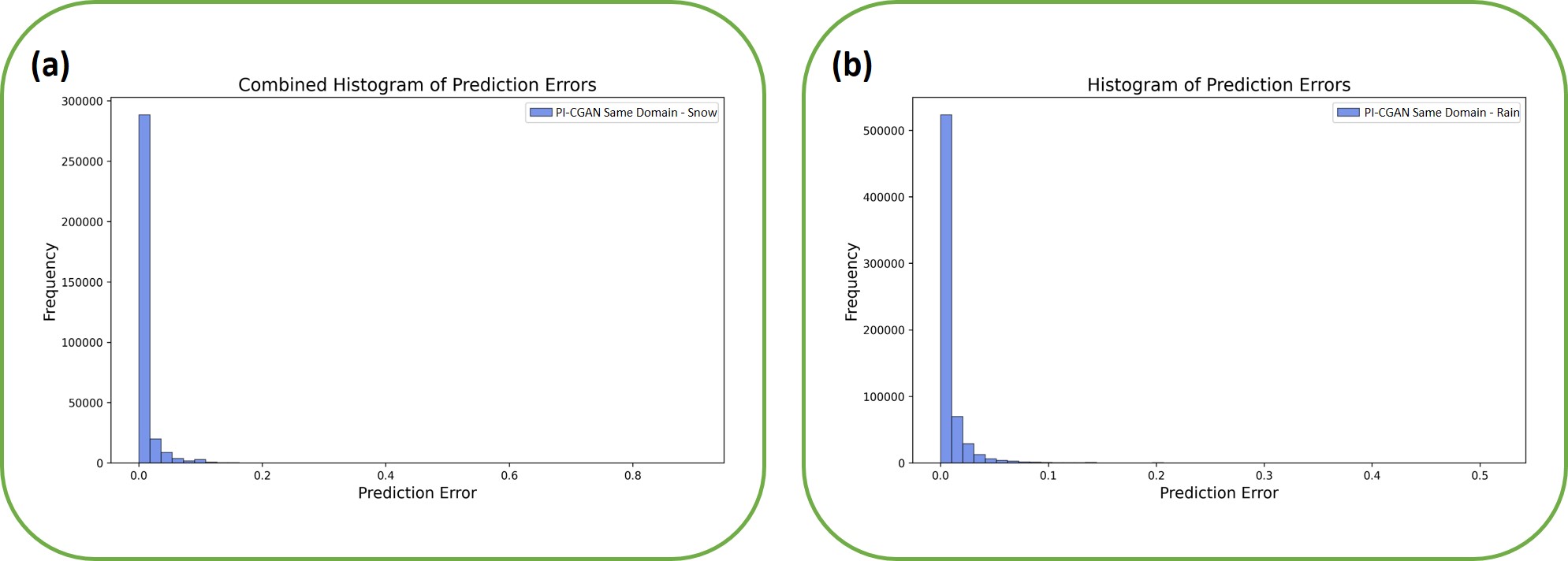}}
\caption{\textbf{Error Histogram} of the prediction errors between the Ground Truth and our PICWGAN Generated Intensity for Same-Domain LiDAR Intensity Transfer on a) Snow Weather CADC dataset and b) Rain Weather Boreas dataset }
\label{fig-eh}
\end{figure*}

    The error histogram, shown in Fig. \ref{fig-eh}, illustrates the difference between the generated and ground-truth intensity values. For both snow and rain, the narrow and symmetric error distribution centered around zero suggests minimal deviation and low frequency of large errors, highlighting the model's precision in producing accurate intensity values. This metric quantifies per-pixel performance, ensuring most intensity values are correctly predicted across varying conditions.

\subsubsection{\textbf{Mean Squared Error (MSE)}}
    The Mean Squared Error (MSE) quantifies the overall discrepancy between the generated and ground-truth intensity values (ranging from 0 to 1). Expressed as a percentage of the maximum possible deviation (i.e., 1), the MSE values for snow (0.0010) and rain (0.0003) correspond to 0.10\% and 0.03\%, respectively, indicating minimal deviation. These results emphasize the model’s precision in replicating intensity values and its robustness in handling challenging weather conditions.

\subsubsection{\textbf{Structural Similarity Index (SSIM)}}

    The Structural Similarity Index (SSIM) quantifies how well the generated intensity images replicate the structural characteristics of the ground-truth images. High SSIM scores, specifically 0.78 (with a standard deviation of 0.0068) for snow and 0.83 (with a standard deviation of 0.0092) for rain, indicate that the model accurately captures the spatial patterns and distribution of intensity values, which are critical for simulating LiDAR intensity data. The low standard deviation values highlight the model's consistency in producing reliable results across varying samples. Preserving the structural fidelity of LiDAR intensity is essential for downstream tasks, as it ensures accurate modeling of surface properties and material reflectivity, even in adverse weather conditions.

\subsubsection{\textbf{Kullback-Leibler (KL) Divergence}}
    
   The Kullback-Leibler (KL) divergence measures how much the probability distribution of the generated intensity values deviates from that of the ground-truth data. Lower KL divergence values indicate a closer match, with our model achieving 0.2726 for snow and 0.1155 for rain, demonstrating strong alignment with real-world distributions. In contrast, the KL divergence between the intensity distribution of the CADC dataset (snow) and the VoxelScape physics-only intensity is significantly higher at 10.48, while between the Boreas dataset (rain) and VoxelScape physics-only intensity, it is 10.30. This stark difference highlights the limitations of purely physics-based simulations and underscores the effectiveness of our model in generating realistic LiDAR intensity distributions that statistically align with real-world conditions.

\subsubsection{\textbf{Wasserstein Distance}} 
    The Wasserstein distance provides a geometric measure of the discrepancy between the generated and real-world intensity distributions. Low Wasserstein distance values, specifically 0.0044 for snow and 0.0064 for rain, confirm the close alignment of the generated intensities with real-world data. This metric complements KL divergence by offering an interpretable measure of how separated the two distributions are from each other. The minimal distances highlight the model's robustness and accuracy in simulating LiDAR intensity data across different precipitation types, ensuring reliable performance under diverse weather conditions.

\subsection{\textbf{Downstream Task: 3D Object Detection}}


We trained PV-RCNN, a state-of-the-art 3D object detector \cite{pvrcnn_shi2020pv} using OpenPCDet \cite{openpcdet2020} on various data combinations as shown in Table \ref{tab:data_comb} and evaluated its performance on real-world adverse weather datasets to assess the impact of our PICWGAN-generated intensities for real-world applications. Specifically, the CADC dataset was used for snow conditions, while KITTI-Rain was used for rain conditions. The evaluation was conducted on 2,500 frames of real-world datasets (CADC for snow and KITTI-Rain for rain) to ensure statistically significant results. The results demonstrate how the generated intensities enhance the realism of simulated data and improve model generalization.

\begin{table}[htbp]
\centering
\caption{Training Data Combinations for Snow and Rain Conditions}
\label{tab:training_data}
\resizebox{\columnwidth}{!}{%
\begin{tabular}{llll}
\toprule
\textbf{Condition} & \textbf{Method} & \textbf{Data Combination} & \textbf{Total Frames} \\
\midrule
\multirow{5}{*}{\textbf{Snow}} 
  & \multirow{3}{*}{\textbf{Baseline}} 
      & VS-Phy & 4000 \\
  & 
      & VS-Phy + CADC & 2000+2000 \\
  & 
      & CADC & 4000  \\
\cline{2-4}
  & \multirow{2}{*}{\textbf{Proposed}} 
      & VS-PICWGAN & 4000 \\
  & 
      & VS-PICWGAN + CADC & 2000+2000 \\
\midrule
\multirow{5}{*}{\textbf{Rain}} 
  & \multirow{3}{*}{\textbf{Baseline}} 
      & VS-Phy & 4000 \\
  & 
      & VS-Phy + KITTI-Rain & 2000+2000 \\
  & 
      & KITTI-Rain & 4000  \\
\cline{2-4}
  & \multirow{2}{*}{\textbf{Proposed}} 
      & VS-PICWGAN & 4000 \\
  & 
      & VS-PICWGAN + KITTI-Rain & 2000+2000\\
\bottomrule
\end{tabular}%
}
\label{tab:data_comb}
\end{table}

The nomenclature used in Table \ref{tab:data_comb} and Table \ref{tab:od-snow-rain} is as follows: VS-Phy refers to VoxelScape data with physics-based simulated intensities under snow or rain conditions, while VS-PICWGAN denotes VoxelScape data enhanced with PICWGAN-generated intensities under snow or rain conditions.

Note: For rain, the Boreas dataset, used for intensity simulation, lacks ground truth 3D object labels, making it unsuitable for the downstream task. Furthermore, no publicly available labeled rain-weather dataset with extensive frames currently exists. To address this, we used the simulated KITTI-Rain dataset, which augments the KITTI dataset with rain effects, for object detection experiments. The PICWGAN was trained on KITTI-Rain data to enable valid comparisons for rain conditions and ensure a consistent evaluation framework across both snow and rain scenarios.

\begin{table*}[ht]
\centering
\renewcommand{\arraystretch}{2} 
\resizebox{\linewidth}{!}{ 
\begin{tabular}{llcccccccccccc}
\toprule
\textbf{Class} & \textbf{Data (Snow)} & \multicolumn{4}{c}{\textbf{Snow}} & \textbf{Data (Rain)} & \multicolumn{4}{c}{\textbf{Rain}} \\
\cmidrule(lr){3-6} \cmidrule(lr){8-11}
& & \textbf{3D (IoU - 0.7)} & \textbf{3D (IoU - 0.5)} & \textbf{BEV (IoU - 0.7)} & \textbf{BEV (IoU - 0.5)} & & \textbf{3D (IoU - 0.7)} & \textbf{3D (IoU - 0.5)} & \textbf{BEV (IoU - 0.7)} & \textbf{BEV (IoU - 0.5)} \\
\midrule
Car & VS-Phy               & 4.08   & 10.58   & 4.42   & 12.42   & VS-Phy               & 18.89  & 33.67  & 20.45  & 36.34 \\
    & VS-PICWGAN                & 5.98   & 14.98   & 6.60   & 16.60   & VS-PICWGAN                & 24.23  & 37.89  & 26.78  & 40.78 \\
    & VS-Phy + CADC        & 30.35  & 38.35   & 31.68  & 44.68   & VS-Phy + KITTI-Rain  & 43.67  & 62.34  & 46.23  & 64.38 \\
    & VS-PICWGAN + CADC         & 37.73  & 42.73   & 38.04  & 47.04   & VS-PICWGAN + KITTI-Rain   & 46.89  & 65.78  & 49.43  & 68.34 \\
    & CADC [4000 Frames]     & 45.18  & 50.98   & 47.60  & 53.60   & KITTI-Rain [4000 Frames] & 48.90  & 69.78  & 51.34  & 72.34 \\
    & CADC [2000 Frames]     & 29.58  & 36.78   & 31.10  & 39.50   & KITTI-Rain [2000 Frames] & 42.12  & 59.45  & 41.56  & 62.42 \\
\midrule
Pedestrian & VS-Phy         & 9.64   & 15.54   & 10.39  & 16.39   & VS-Phy               & 13.78  & 21.23  & 15.12  & 23.74 \\
           & VS-PICWGAN          & 10.06  & 21.06   & 10.64  & 22.64   & VS-PICWGAN                & 15.89  & 24.23  & 17.34  & 27.89 \\
           & VS-Phy + CADC  & 28.12  & 34.02   & 28.22  & 38.22   & VS-Phy + KITTI-Rain  & 32.10  & 49.67  & 35.67  & 52.34 \\
           & VS-PICWGAN + CADC   & 33.57  & 38.57   & 34.80  & 44.80   & VS-PICWGAN + KITTI-Rain   & 35.12  & 51.12  & 37.89  & 55.34 \\
           & CADC [4000 Frames] & 40.06  & 45.06   & 41.64  & 48.64   & KITTI-Rain [4000 Frames] & 37.56  & 57.45  & 40.12  & 60.12 \\
           & CADC [2000 Frames] & 25.36  & 29.36   & 27.94  & 32.94   & KITTI-Rain [2000 Frames] & 30.45  & 46.89  & 33.78  & 49.34 \\
\midrule
Truck & VS-Phy             & 2.03   & 7.23    & 3.04   & 8.54    & VS-Phy               & 9.45   & 16.12  & 10.89  & 18.67 \\
 (For Snow)     & VS-PICWGAN              & 4.06   & 8.46    & 4.62   & 11.52   & VS-PICWGAN   & 13.12  & 19.12  & 14.78  & 23.34 \\
Cyclist  & VS-Phy + CADC      & 28.28  & 34.28   & 29.39  & 38.39   & VS-Phy + KITTI-Rain  & 29.06  & 45.89  & 31.45  & 48.45 \\
 (For Rain)     & VS-PICWGAN + CADC       & 31.22  & 36.22   & 33.44  & 37.44   & VS-PICWGAN + KITTI-Rain   & 30.65  & 48.89  & 33.12  & 50.78 \\
      & CADC [4000 Frames]   & 41.76  & 45.46   & 44.52  & 49.02   & KITTI-Rain [4000 Frames] & 32.78  & 52.78  & 35.34  & 55.34 \\
      & CADC [2000 Frames]   & 21.22  & 28.22   & 25.44  & 31.44   & KITTI-Rain [2000 Frames] & 27.23  & 41.12  & 28.89  & 43.45 \\
\bottomrule
\end{tabular}
}
\caption{Performance comparison of PC-RCNN model trained on different dataset combinations with and without Physics-Informed Cycle-Consistent Weather GAN-generated (VS-PICWGAN) intensities for snow and rain, evaluated using BEV and 3D IoU at 0.5 and 0.7 thresholds, where higher IoU values indicate stricter matching criteria.}
\label{tab:od-snow-rain}
\end{table*}


The results, summarized in Table \ref{tab:od-snow-rain}, show that the use of PICWGAN intensities consistently improved 3D object detection performance compared to the baseline methods. For both snow and rain conditions, models trained on VoxelScape data with PICWGAN intensity outperformed those trained on the VoxelScape data with physics-based simulated intensities in adverse weather conditions, demonstrating the PICWGAN's effectiveness in enhancing the realism of simulated data.


The best performance was observed when combining VoxelScape (PICWGAN Intensity) data with real-world datasets (CADC for snow and KITTI-Rain for rain), outperforming all other data combinations, except models trained solely on real-world data. This outcome is expected, as models trained on real-world data generally perform better when tested in the same domain, due to the alignment of the domain-specific terrain, object distribution, and intensity patterns with the test dataset. However, it is significant that models trained on the combination of VoxelScape (PICWGAN Intensity) and real-world datasets performed comparably to those trained exclusively on real-world data. These results confirm that the PICWGAN generates realistic intensities that closely mimic real-world conditions. By aligning simulated and real intensity distributions, it bridges the sim-to-real gap, improving model generalization across varying terrains and objects.


\section{Conclusion}

This work presents a physics-informed PICWGAN framework for simulating LiDAR under adverse weather conditions. By integrating physics-based modeling with learning-based translation, the approach captures both intensity variations and geometry-consistent degradations induced by rain and snow, effectively bridging the simulation-to-reality gap. Qualitative and quantitative evaluations demonstrate the model’s ability to reproduce key weather effects such as attenuation and scattering, yielding realistic LiDAR intensity distributions. Metrics including MSE, SSIM, KL divergence, and Wasserstein distance show strong agreement between generated and real data, confirming both structural and statistical fidelity. Furthermore, in downstream 3D object detection tasks, the simulated data improves model performance, highlighting the importance of physically consistent augmentation for robust perception. Overall, this work underscores the value of combining physics-based insights with generative models for realistic simulation of adverse weather. It provides a step toward more reliable perception systems for autonomous driving and robotics, with future directions including more advanced atmospheric modeling, dynamic weather scenarios, and extension to other sensor modalities.

\section*{Acknowledgment}

The authors thank the University of New Brunswick, Canada, for access to the Digital Research Alliance of Canada, the Data Science Research Division of the Frontier and Futuristic Technologies Division, DST, GOI, New Delhi, for their support, and SimDaaS Autonomy Pvt. Ltd. for assistance in carrying out this work.

\bibliographystyle{ieeetr}
\bibliography{reference}
\patchcmd{\thebibliography}{\baselineskip15pt}{\baselineskip 12pt}{}{}

\vspace{-1cm} 

\begin{IEEEbiography}
[{\includegraphics[width=1in,height=1.25in,clip,keepaspectratio]{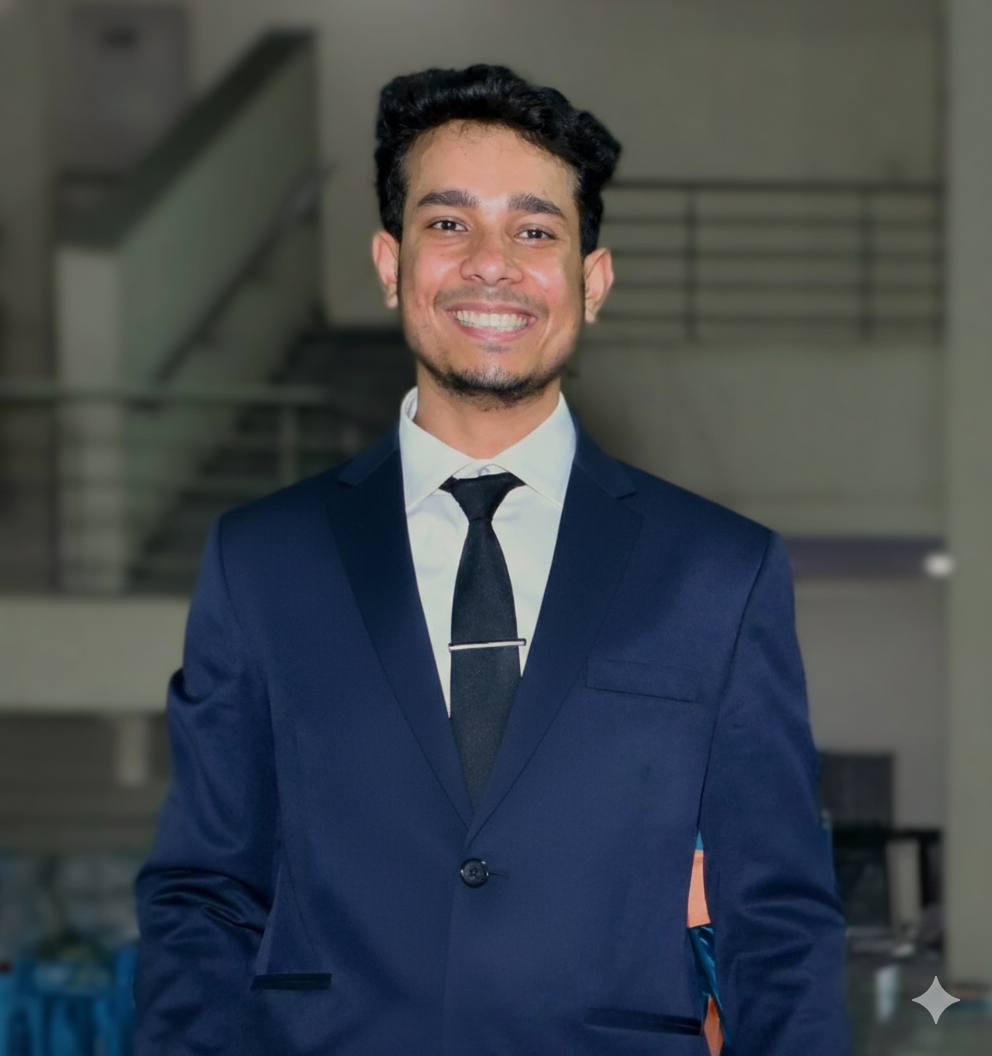}}]{Vivek Anand} completed his B.Tech in Civil Engineering from KIIT University in 2019, M.Tech in Geoinformatics from Delhi Technological University (DTU) in 2021, and PhD in Geoinformatics from the Indian Institute of Technology (IIT), Kanpur in 2026. He is currently a Fellow of Academic and Research Excellence at IIT Kanpur. His research focuses on sim-to-real transfer and computer vision applications in autonomous systems and intelligent transportation systems.
\end{IEEEbiography}

\vspace{-1cm} 

\begin{IEEEbiography}
[{\includegraphics[width=1in,height=1.25in,clip,keepaspectratio]{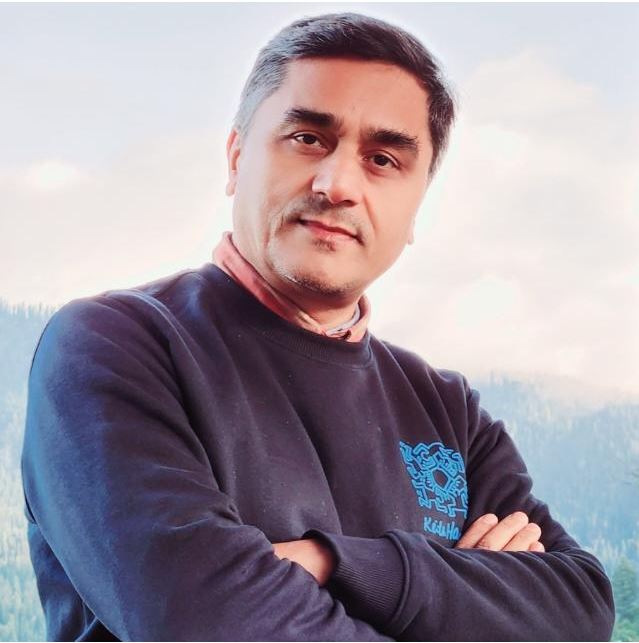}}]{Bharat Lohani} earned his PhD from the University of Reading, the UK, in 1999. He is currently a
Professor at the Department of Civil Engineering,
Indian Institute of Technology, Kanpur, India. Dr. Lohani mainly focuses on the modeling of the physical environment using high-resolution remotely sensed data, with particular emphasis on airborne, drone, mobile, and terrestrial LiDAR data and photographs. His current research interests are in LiDAR Data Classification using Deep Learning Techniques, LiDAR application in Forest and Water Conservation, Solar Insolation Estimation, and especially LiDAR simulation for autonomous systems.
\end{IEEEbiography}

\vspace{-1cm} 

\begin{IEEEbiography}    
[{\includegraphics[width=1in,height=1.25in,clip,keepaspectratio]{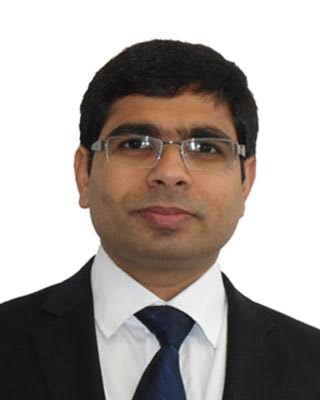}}]{Rakesh Mishra} earned his Ph.D. from the University of New Brunswick, Canada, in 2017. Currently serving as an Adjunct Professor in the Department of Geodesy and Geomatics Engineering at the University of New Brunswick, he also holds the position of Chief Technology Officer at SceneSharp Technologies Inc. in Fredericton, NB. Dr. Mishra has significantly contributed to the field through his authorship or co-authorship of numerous research articles covering topics such as image fusion, object detection, hyperspectral imaging, and LiDAR. His expertise extends to areas such as multi-sensor data fusion, computer vision, hyperspectral imaging, and AI-based object detection utilizing diverse sensor data. Driven by his leadership, he has successfully overseen the development of more than six technologies, which are currently employed in industries related to defense and forestry.
\end{IEEEbiography}

\vspace{-1cm} 

\begin{IEEEbiography}    
[{\includegraphics[width=1in,height=1.25in,clip,keepaspectratio]{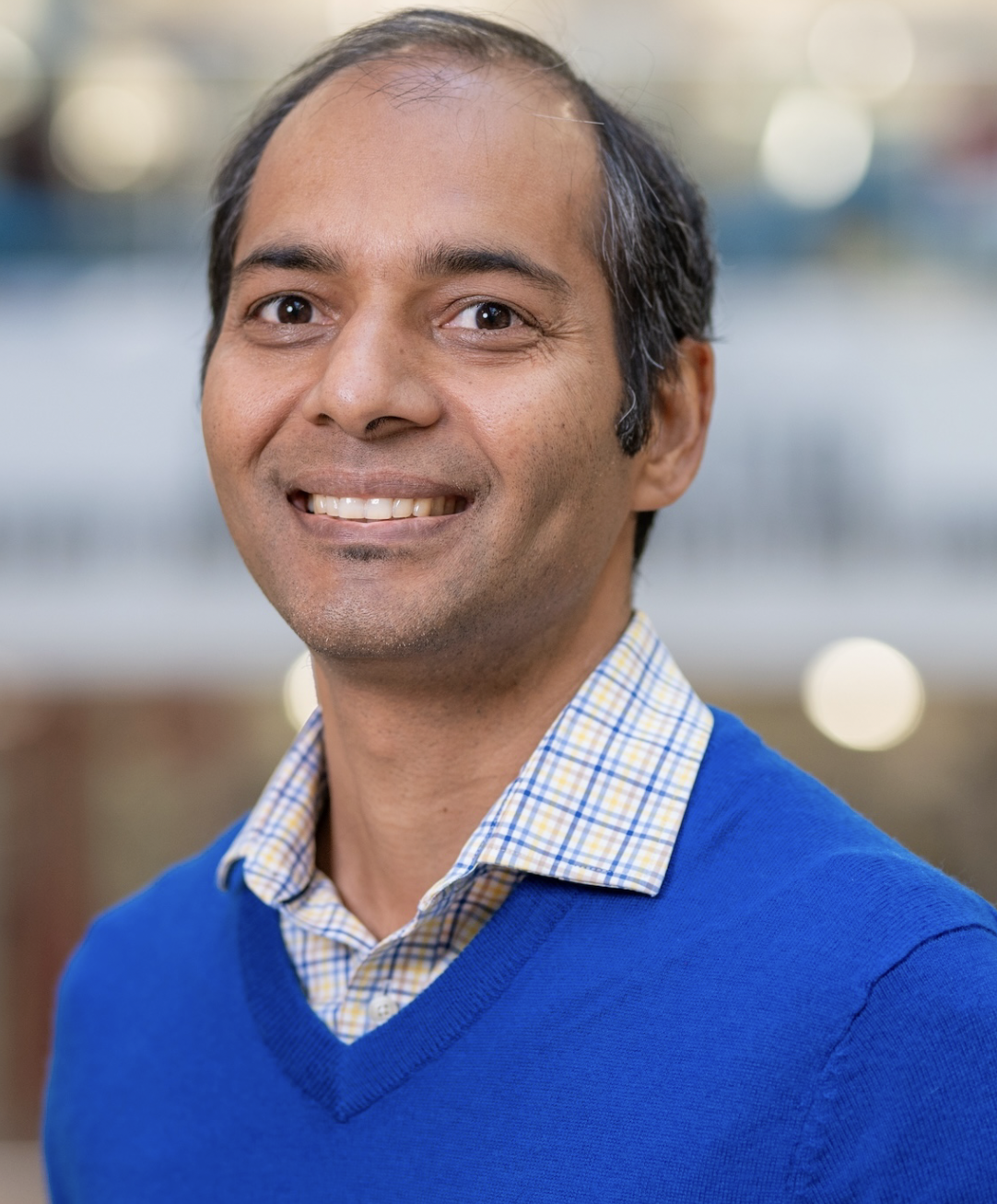}}]{Gaurav Pandey} recieved his B-Tech from IIT Roorkee in 2006 and completed his Ph.D. from the University of Michigan, Ann Arbor in December 2013. He is currently an Associate Professor at the Engineering Technology and Industrial Distribution (ETID) department of Texas A\&M University. He previously worked at Ford Motor Company as a Technical Leader in the automated driving division. Dr. Pandey also served as an assistant professor at the Electrical Engineering Department of the Indian Institute of Technology (IIT) Kanpur from 2015 to 2016. Dr. Pandey is an experienced professional working in the field of autonomous vehicles and robotics, with significant achievements in 3D lidar processing, sensor data fusion, 3D mapping, and localization research.
\end{IEEEbiography}

\end{document}